# Enabling Commercial Autonomous Space Robotic Explorers

Thomas Yuang Li, *student member IEEE*
Shaoshan Liu, *senior member IEEE*

*Abstract:*

*In contrast to manned missions, the application of autonomous robots for space exploration missions decreases the safety concerns of the exploration missions while extending the exploration distance since returning transportation is not necessary for robotics missions.*

*In addition, the employment of robots in these missions also decreases mission complexities and costs because there is no need for onboard life support systems: robots can withstand and operate in harsh conditions, for instance, extreme temperature, pressure, and radiation, where humans cannot survive.*

*In this article, we introduce environments on Mars, review the existing autonomous driving techniques deployed on Earth, as well as explore technologies required to enable future commercial autonomous space robotic explorers. Last but not least, we also present that one of the urgent technical challenges for autonomous space explorers, namely, computing power onboard.*

## 1. Introduction

With the increased societal desire and technological capability to explore outer space, the plots in sci-fi are becoming a reality. Robots for space applications will become common in the near future.

In the past, robots have been sent into space for varied purposes, e.g., taking photographs and performing mineral composition analysis etc.. In contrast to manned missions, the application of autonomous robots for space exploration missions decreases the safety concerns of the exploration missions while extending the exploration distance since returning transportation is not necessary for robotics missions. In addition, the employment of robots in these missions also decreases mission complexities and costs because there is no need for onboard life support systems: robots can withstand and operate in harsh conditions, for instance, extreme temperature, pressure, and radiation, where humans cannot survive.

Most of the space robotic exploration missions today rely on remote control from Earth, however, this method suffers from extremely long communication latencies, thus leading to the lack of efficiency for an operator to receive information, make a decision on how to respond, and issue commands to the spacecraft. To improve the efficiency of robotic exploration missions, there have been several attempts by NASA to enable autonomous robotic navigation on Mars [6].

As shown in Figure 1, we envision a future where commercial autonomous robotic explorers will explore and construct basic infrastructure on Mars, making it habitable for mankind. While there have been a few successful attempts to deliver exploration robots to planets like Mars, how to develop autonomous robots suitable for commercial space exploration missions still requires a lot more research.

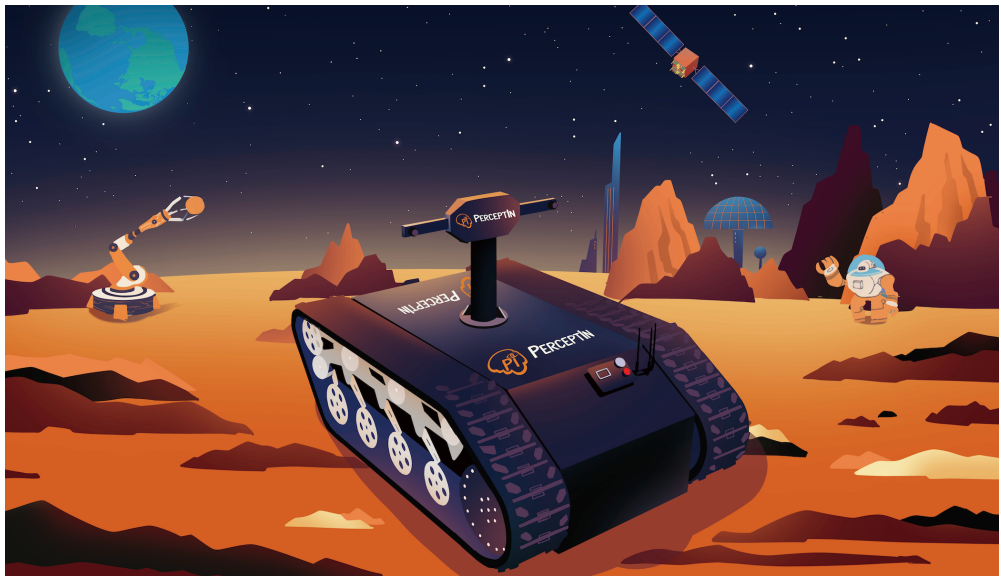

Figure 1: the future of commercial space exploration

In this article, we explore the autonomous driving technologies, including localization, perception, planning and control, *etc.*, required to enable a commercial space exploration robot, as well as how to integrate these technologies into a working system.

The rest of this paper is organized as follows, in section 2 we introduce the environments of Mars, and explore the challenges of enabling autonomous robotic explorers on Mars. In section 3, we review autonomous driving technologies already deployed on Earth. In section 4, we explore technologies required to enable future commercial autonomous space robotic explorers. In section 5, we present one urgent challenge for space exploration mission, namely, computing power. We conclude in section 6.

## 2. Destination Mars

Mars is a potential destination for autonomous space robotic explorers. Currently, the main focus of space exploration is Mars, for it is relatively close to Earth and shares many macro level similarities to Earth, like the existence of atmosphere and the evidence of past flowing water.

The purpose of Mars exploration can be categorized into understanding the evolution of the Martian environment, examining the current conditions of Mars, and search for past, present, and future potential for life. Mars exploration also lays the foundation for potential exploitation of Martian resources, and ultimately, mankind expansion.

Up until now, uncertainties about the Martian environment and associated high cost prevented human exploration of Mars. Therefore, Mars surface exploration is currently accomplished with robot explorers. Explorers collect and send back scientific data on the Martian surface and thus pave the way for future human exploration.

As illustrated in Figure 2, Mars environment is drastically different from Earth, especially in terms of atmospheric composition, temperature, and geologic features. These differences pose potential challenges for robot design.

The Martian atmosphere is 96% carbon dioxide compared to less than 1% on Earth. The temperature on Mars can be as high as 70 degrees Fahrenheit or as low as about -225 degrees Fahrenheit. Because the atmosphere is so thin, heat from the Sun easily escapes this planet, and thus temperatures at different altitudes have tremendous variations. In addition, occasionally, winds on Mars are strong enough to create dust storms that cover much of the planet. After such storms, it can be months before all of the dust settles [2].

The geological features on Mars are generally more extreme than those on Earth. For instance, the deepest canyon on Mars is around 7 km in depth whereas on Earth, it is 1.8 km [1]. Martian soil is also different from soil on Earth, it is composed of fine regolith, or unconsolidated rock powder, which lacks traction and also affects visibility during Martian storms and that has huge implications for sensors used in autonomous movement.

The extreme geological features combined with low traction soil and low gravity pose challenging design questions for autonomous explorers, in terms of both hardware and software.

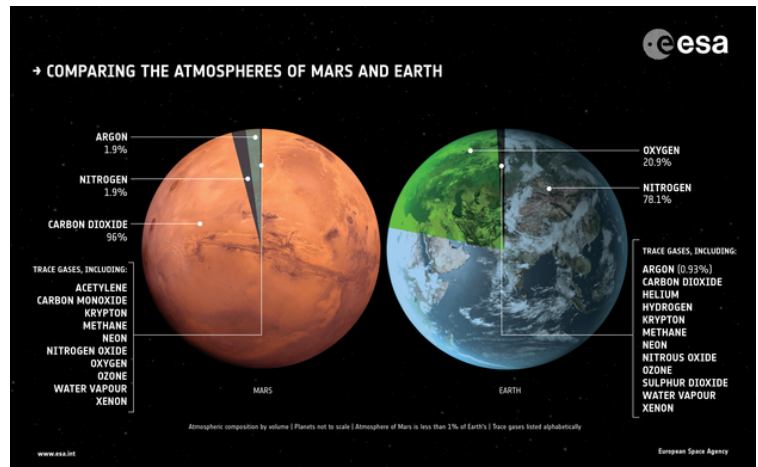
Figure 2: atmosphere comparisons between Earth and Mars (source: [2])

## 3. Autonomous Driving on Earth

Before examining the autonomous driving technologies on space explorers, we shall first explore the autonomous driving technologies used on Earth with a focus on sensing, perception, and decision [3,4]. Figure 3 shows the autonomous driving technology architecture deployed on Earth. Note that for space robotic exploration missions, the technology architecture is similar to the one presented in Figure 3 but with variations to adapt to the destination's environment. For instance, on Mars there is no GPS system deployed and thus we cannot rely on GPS systems for robotic localization.

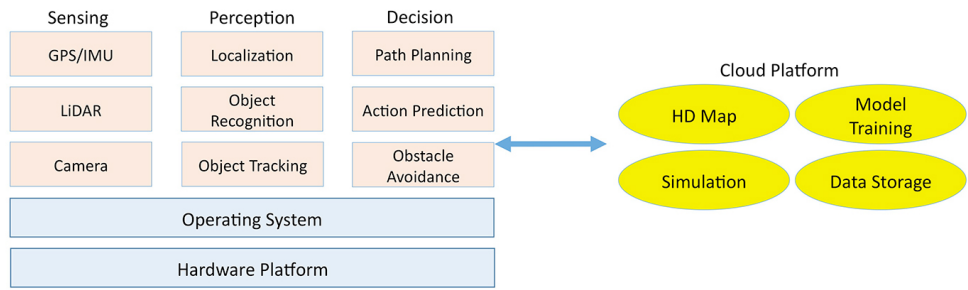

Figure 3: autonomous driving technology architecture

## 3.1 Sensing

Sensing, in essence, is using sensors to collect data of the environment. There are several types of sensors that are commonly used on Earth. Each type has advantages and drawbacks, so multiple types of sensor are usually used together to ensure the reliability of sensing. Four main types are:

1. **GPS/IMU**: The GPS/IMU system provides information on vehicle position in regard to the world. While GPS provides relatively accurate position information, it updates at a low frequency around 10Hz, hindering position accuracy once the vehicle is traveling at a fast speed. An inertial measurement unit (IMU) is an electronic device that measures and reports a body's specific force, angular rate, and sometimes the orientation of the body, using a combination of accelerometers, gyroscopes, and sometimes magnetometers. The addition of IMU fills the gap between each GPS position update since IMU updates at a much higher frequency, around 200 Hz.

2. **LiDAR**: LiDAR, which stands for Light Detection and Ranging, works by shooting out beams of laser and measuring the returning time and frequency, and thus measuring the distance of each point. As a result, it produces high definition, with-depth images composed of a group of points. LiDAR is praised for its high accuracy.

3. **Camera**: Cameras capture accurate visual and other light spectrum (e.g. infrared) representations of the surrounding environment. Cameras can detect features in the environment and track them. Cameras update at a sufficient speed and is more economical compared to LiDAR. However, cameras rely on the lighting of the environment.

4. **Radar and Sonar**: Radar and sonar are generally used to detect close objects. Radar transmits radio waves and captures the waves' reflections from objects. Using this method, a radar system can detect the range, angle, and velocity of the object based on the time it takes for waves to return and the returning frequency change. Sonar works similarly to radar, except for it transmits high frequency sound waves that exceeds the human hearing range rather than radio waves. Using sound waves, sonar system is able to detect objects such as clear plastic that are difficult to detect using radar. Radar and sonar usually serve as hazard avoidance devices.

## 3.2 Perception

Perception is the analysis of the raw data collected by the sensors. Perception systems make sense of and utilize raw data. On earth, perception is mainly concerned with understanding of the environment as well as localization.

GPS/IMU system is used for localization. GPS provides relatively accurate position information but updates at a lower frequency. IMU updates at a higher rate but over time accumulates error. The advantages of both GPS and IMU can be utilized when combining their data via a Kalman filter. However, the drawbacks of this system are obvious. Even at its optimal performance, the system has an accuracy of only around one meter, which is not sufficient for the complex road environment on Earth. Although GPS is a reliable localization method on Earth, since there is no GPS system deployed on Mars, robotic explorers have to rely on cameras and other means for accurate localization.

Cameras can also be used for localization. By using stereo cameras, the image pairs generated can be used for triangulation, and thus constructing 3D vision. Localization using this method is achieved by comparing the salient features in the 3D image collected with those in existing high-definition maps. However, camera vision is prone to inaccuracy due to inconsistency in natural lighting conditions which cameras rely on.

LiDAR is a relatively reliable and accurate means of perception. The application of LiDAR effectively increases the accuracy of localization in an urban environment to around 10 cm, ten times that of GPS/IMU. The only downside of LiDAR is that when there are excessive particles suspended in the air, such as particulate matters and raindrops, the measurement generated will become noisy.

## 3.3 Decision

The decision stage involves the combination of information obtained in the perception stage and ultimately using this information to reach the destination. This stage can be subdivided into path planning and obstacle avoidance.

Path planning on earth involves two parts, behavioral prediction of maneuverable objects, such as other cars, around the autonomous vehicle, and the behavior planning of the vehicle itself [5]. In order to ensure safety while interacting with humans and other vehicles when driving, autonomous vehicles need to predict the possible action of others so the vehicle can take action accordingly. Such prediction is created by first, finding reachable positions of other moving objects, and then, utilizing a probability distribution to find the position of highest probability. Path planning is then achieved using various methods which take into account a wide range of factors, including safety, comfort, and efficiency to generate the optimal path.

Obstacle avoidance concerns the safety of the vehicle. It can be divided into two layers, which are proactive avoidance and reactive avoidance. On Earth, proactive avoidance is based on the prediction of traffic. Specifics include time-to-collision, predicted minimum distance, etc. These measurements are then used in the local path re-planning to ensure the vehicle's safety. Reactive avoidance serves as the last line of defense. If the reactive sensors, such as radar and sonar, detect an obstacle in the path, they override the current tasks and trigger an emergency response like braking.

## 4. Mars Explorer Autonomy

The above introduced technologies and methods work well on Earth. However, when it comes to space explorers, the environment in which the vehicle operates in is vastly different from the environment on Earth. As a result, autonomous driving and navigation methods also have to be modified.

On planets other than Earth, there is no set traffic system nor other maneuverable objects, which simplify the autonomous driving system in the sense that explorers do not have to localize at a centimeter level precision and it can maneuver without any restriction and plan path without concern regarding crashing into another vehicle. However, the underdeveloped infrastructure in space also poses challenges on the design of autonomous driving systems.

In space, there is no available GPS system or detailed surface map. The terrain condition is also much more complex and requires more considerations compared to driving on Earth, for there is no road infrastructure in space.

We shall now examine the autonomous driving technologies and challenges for Mars explorers also with a focus on perception and decision. This environment is particularly challenging for localization and path planning of the robot explorers.

## 4.1 Perception: Localization

Without the availability of GPS and known map, localization for Mars explorer robots is keeping track of its motion trajectory while observing the surrounding environment and estimating its position in the environment. Mars explorer localization rely on the camera and IMU for localization. One additional sensor that has been applied to Mars explorers is a star tracker, an optical device that measures the position of stars. There are mainly three methods for localization: vision-based simultaneous localization and mapping (SLAM), dead reckoning based on IMU, and star tracking.

The lack of detailed maps of Mars leaves the robot explorer having to navigate in an unknown environment while keeping track of its path. SLAM is a method of robot constructing the map of the environment and localizing with regard to its surroundings simultaneously. While driving, an explorer can estimate its trajectory with a structure-from-motion algorithm. This trajectory is then used to create a layout of the environment by incorporating matching and triangulation. This constructed environment layout can then in turn be used for further localization with regard to the environment [6].

Specifically, visual odometry is a popular SLAM technique. Visual odometry is the estimation of the motion of explorer using camera vision. It first detects the salient features on the image and then estimates the 3D positions of selected features by stereo matching. It tracks the salient features through a sequence of optical images and thus determines the change of positions of the explorer. Note that visual odometry has already been applied on the NASA Mars Exploration Mission in 2003 with promising results.

The second method to achieve localization on Mars is dead-reckoning, the process of estimating explorer's current position using its previous position and updating based on estimated information such as the velocity. IMU and wheel encoder are commonly used in dead reckoning [7]. By incorporating gyroscope and accelerometer, IMU estimates the linear acceleration and rotation rate of the explorer. Then through integration over time, these measurements are used for dead reckoning and pose estimation. However, since dead reckoning relies on integration over time, the accuracy of such method also decreases over time and thus cannot be used for effective localization in long-distance operations. Likewise, another dead reckoning technique, wheel odometry, also has drawbacks when operating on Mars. Since the surface of Mars is covered with fine regolith, and since the explorer often has to pass through rugged terrain, wheel slip is common, which makes wheel odometry inaccurate.

Star tracking is another localization method available for use during planetary exploration [8]. On Mars, the lack of GPS system and global magnetic field rise challenge for determining the orientation of the explorer. Using star tracking method, a camera-based star tracker identifies the position of a known set of stars and then compares them with the absolute known position of stars stored in memory. As many star positions have been measured to high accuracy, star tracking allows the explorer to determine its orientation. However, star trackers would only work at night and may fail when used in bright environments, and thus this method fails to enable accurate localization at all times.

## 4.2 Perception: Hazard Avoidance

The complex terrain on Mars makes terrain assessment a crucial component of autonomous navigation. Similar to the methods applied in autonomous driving on Earth, Mars Explorers use an obstacle avoidance system that is divided into two layers, proactive and reactive.

In the proactive layer, the 3D point cloud generated by stereo vision and triangulation constructs the shape of the environment and thus detects obstacles and hazards. With robust and efficient on-board processors and the reliable algorithm for perception of hazards of current processors, it is fair to say that the challenge for terrain assessment does not lie in the geometry of the terrain, *e.g.* rocks, but rather in non-geometric aspects where hazards are not obvious for detection, such as the load-bearing properties of the terrain.

In the reactive layer, basic level information like current vehicle tilt and the feedback from wheels and suspension are taken into consideration. As the last line of defense, if any feedback from sensors is abnormal, the explorer's emergency response will be immediately triggered. For instance, the discrepancy between wheel odometry and visual odometry exceeds a set limit, the vehicle can conclude that it is experiencing severe wheel slippage and thus generate an alternative path.

## 4.3 Path Planning

In order to navigate from one point to another, the explorer generates a series of direction oriented waypoints, leading up to the goal destination. Each waypoint is reached by repeating the process of terrain assessment and path selection.

The basic hazard avoidance capabilities discussed above are sufficient to stop a vehicle once it is in a risky situation. However, in order to achieve efficiency and to increase safety, terrain assessment is also incorporated into the path planning process.

Various methods have been applied for explorer path planning. One such method transforms the environment information gained from analyzing stereo images into grid cells, which cover the area surrounding the explorer and thus building a local traversability map stored in the explorer's memory system.

This map is centered around the explorer and is updated constantly as the explorer moves and gains new terrain assessment information. Then, plane fitting is applied to the traversability map to assess how safe the explorer will be at each point in the map. The plane models the explorer body, and it is roughly the size of the explorer plus an additional margin for safety.

Centered in every grid cell, a set of 3D points representing the explorer plane are sampled from the point cloud generated by stereo image. The 3D data is analyzed for information regarding the traversability of the explorer, such as tilt and roughness of the terrain. If at a grid cell the explorer plane has excessive tilt, too much residual (indicating that the underlying terrain is too rough), or deviations from the best fit (greater than explorer clearance), the grid cell is then marked as impassable.

In an ideal condition, the explorer plane lays flat on the surface. Based on the above-discussed assessment process, each grid cell is assigned a value that serves as a safety index, reflecting the terrain safety there. As shown in Figure 4, this method can be visualized as the difference in colors of grid cells reflecting a safety index of the environment which is used in path selection.

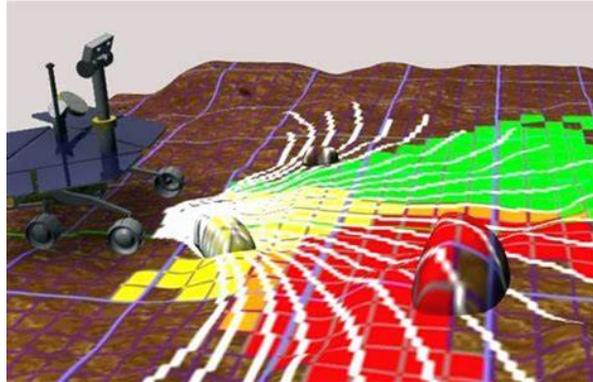

Figure 4: grid-based traversability map (source: https://www-robotics.jpl.nasa.gov/projects/projectVideo.cfm?Project=1&Video=70)

Path selection for explorers follows two principles: safety and efficiency. When candidate motion paths are generated, they are projected onto the traversability map. Each path is assigned an overall safety or traversability evaluation based on the safety index given to the individual grid-cells in the path. Paths are also evaluated for efficiency. The one directly leading to the waypoint is preferred. The efficiency is reduced as the candidate paths deviate from the direct path. By combining safety and efficiency evaluation, the explorer can ultimately select the optimal path to reach the goal destination.

## 4.4 Mars 2020 Explorer

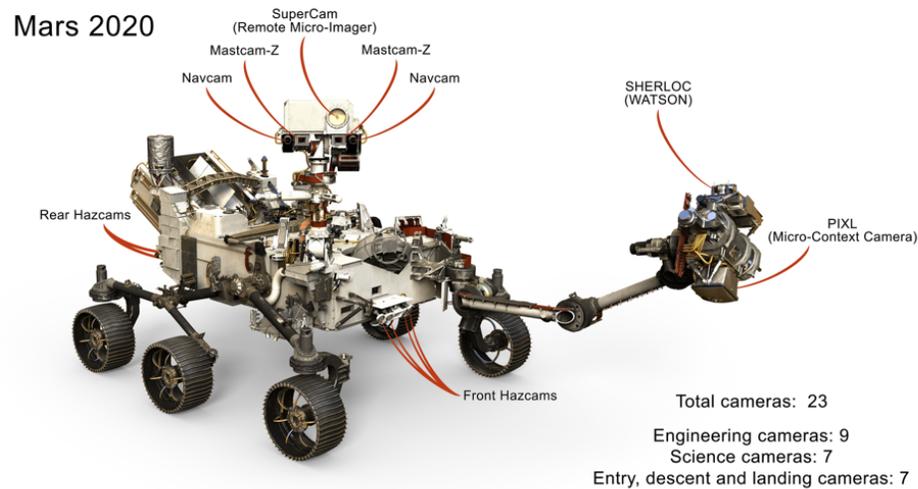

Figure 5: Mars 2020 Explorer (source: https://mars.nasa.gov/resources/21365/mars-2020-cameras/)

The latest Mars explorer project is Mars 2020 by NASA, which is set to launch in July 2020. The new explorer relies solely on camera based systems for navigation. As shown in Figure 5, it is equipped with 23 different cameras responsible for autonomous driving, including six hazard avoidance cameras and one pair of navigation camera.

With one pair mounted in the back and two pairs mounted in the front, the hazard avoidance cameras monitor the surrounding environment of the explorer. They take stereo images and assess the terrain traversability. Hazcams have a broad field of view of 120°. The navigation cameras have a field of view of 45° and are mounted on top of the mast. They also take stereo images but are responsible for macro-level observation.

Compared to the previously sent explorer Curiosity, Mars 2020 has a major upgrade in image processing power, increasing the speed for stereo image and visual odometry calculations. As a result, the explorer will be able to operate at a higher speed, around 152 meters per hour, outperforming its predecessors.

# 5. Challenge: Onboard Computing Capability

In multiple space exploration missions, NASA has demonstrated autonomous rover capabilities. As detailed in [9], autonomous navigation not only improved target approach efficiency, it also proved crucial to maintaining vehicle safety. However, on-board autonomy is often constrained by processor power because of the number of sensor inputs that have to be evaluated in real-time.

For instance, the Opportunity rover is equipped with a IBM RAD6000, a radiation-hardened single board computer, based on the IBM RISC Single Chip CPU [9]. For reliability reasons, the RAD6000 on Opportunity rover was implemented on a radiation-hardened FPGA and runs only at 20 MHz, delivering only over 22 MIPS (million instructions per second). In comparison, a commercial Intel Core i7 can easily deliver 50,000 MIPS running at 2.7 GHz.

Unfortunately, we cannot use these powerful commercial CPUs or GPUs for space exploration missions since these fabrics are not radiation-hardened and thus cannot operate in the harsh environments in space or on other planets. Nonetheless, a few FPGA manufacturers provide radiation-hardened and space-ready FPGA substrates and we can optimize autonomous navigation workloads on these substrates to enable future commercial space exploration missions.

Due to the computing power constraints, only limited autonomous navigation capabilities were turned on in the Opportunity rover. For instance, only low-resolution low-throughput cameras were used for localization, and each update of rover location would take a nontrivial amount of time, up to three minutes for a single location update, as opposed to 30 updates per second for autonomous vehicles on Earth, thus leading to the average speed of rover autonomous navigation under 0.1 miles per hour.

To make matters worse, in order for the Opportunity rover to make long-distance autonomous navigation, ground truth data (*e.g.* 3D reconstruction of the Mars surface for rover localization) on Mars was needed but missing, and this limited the distance the Opportunity rover could travel.

To generate the needed ground-truth data for the Opportunity rover, the rover needs to collect large amount of images on Mars and then perform surface reconstruction or map optimization, an extremely computationally expensive step. As discussed in [10], ground-truth data reconstruction is achieved through employing structure-from-motion algorithms. However, processing structure-from-motion algorithms is beyond the current capabilities of a explorer's on-board computer. To provide an idea of how much computing power it consumes: to reconstruct the city of Rome using 150,000 images would require a 500-node cluster running fully for 24 hours [11].

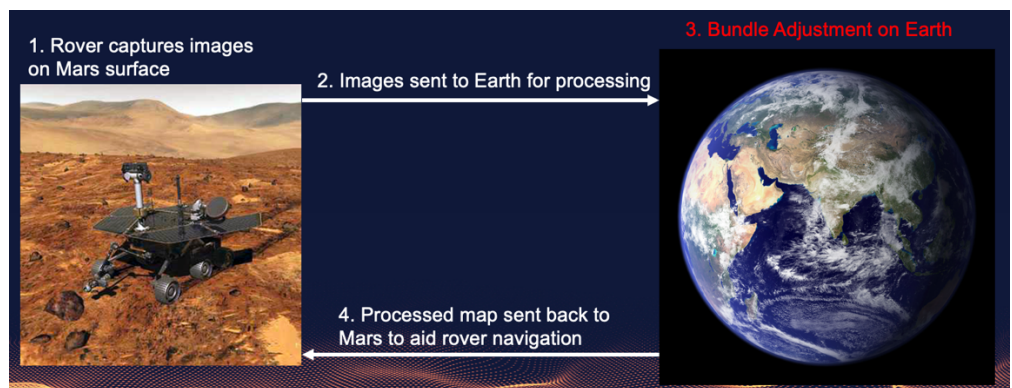
Figure 6: Mars navigation map generation

As shown in Figure 6, what happened on the Opportunity rover was that the rover would run for a very short distance (e.g. a few meters) and collect images, then send the images back to Earth for further processing. Then the rover would then wait for the reconstructed surface map to be transmitted back before it could take the next action.

This is extremely inefficient, as a one-way communication to Mars has a delay up to 20 minutes, depending on the orbital alignment of the planets. In addition, communication is available only twice per sol (Martian day) due to the limited spacecraft around Mars.

We believe the solution lies in developing better computing systems for space explorers. it is thus imperative to implement and optimize autonomous navigation algorithms on radiation-hardened FPGA systems. One example of radiation-hardened FPGA is the Xilinx Virtex-5QV product line.

One such example is presented in [10]: bundle adjustment (BA) is the most computationally demanding step in 3D scene reconstruction, and the authors developed

BA accelerator engine on an embedded FPGA, and experimental results confirm that this design outperformed ARM processors by 50x while maintaining similar power consumption. As more autonomous navigation workloads are implemented and optimized on radiation-hardened FPGAs, we look forward to enable more autonomous navigation scenarios for robotic space explorers.

# 6. Conclusion

Autonomous space exploration robots are still in their infancy. Many challenges have yet to be solved. Progresses in these areas will definitely increase the efficiency of autonomous explorers. The development of autonomous space explorers also opens up exciting possibilities including commercial use, such as exploiting mineral resources on other planets. Eventually, hopefully in the near future, we will have commercial autonomous space robotic explorers building infrastructures for human settlement on Mars.

In this paper, we have introduced environments on Mars, reviewed the existing autonomous driving techniques, as well as explored technologies required to enable future commercial autonomous space robotic explorers. Last but not least, we indicated that one of the urgent technical challenges for autonomous space explorers is computing power: if we had sufficient computing power onboard, we would be able to enable much more efficient robots on Mars. Together, let us build autonomous robots for the space exploration age.

## Authors:

- Thomas Yuang Li is currently a research intern at PerceptIn, his research focuses on building autonomous space explorers for future commercial robotic space exploration mission. Thomas is a student member of IEEE and he can be reached at yuang.li@perceptin.io

- Dr. Shaoshan Liu is founder and CEO of PerceptIn, his research interests include robotics, autonomous driving, SLAM, and edge computing systems for autonomous vehicles. Dr. Liu is a senior member of IEEE and he can be reached at shaoshan.liu@perceptin.io